\documentclass[]{spie}  

\usepackage{macro}

\usepackage{amsmath,amsfonts,amssymb}
\usepackage{graphicx}
\usepackage[colorlinks=true, allcolors=blue]{hyperref}

\title{AutoColor: learned light power control for multi-color holograms}

\author[a]{Yicheng Zhan}
\author[a,b]{Koray Kavaklı}
\author[b]{Hakan Urey}
\author[c]{Qi Sun}
\author[a,*]{Kaan Akşit}

\affil[a]{University College London, Department of Computer Science, London, United Kingdom}
\affil[b]{Koç University, Department of Electrical and Electronics Engineering, Istanbul, Türkiye}
\affil[c]{New York University, Department of Computer Science and Engineering, New York, United States of America}

\authorinfo{Further author information: (Send correspondence to Kaan Akşit)\\Kaan Akşit: E-mail: k.aksit@ucl.ac.uk}

\begin{document} 
\maketitle

\begin{abstract}
Multi-color holograms rely on simultaneous illumination from multiple light sources.
These multi-color holograms could utilize light sources better than conventional single-color holograms and can improve the dynamic range of holographic displays.
In this letter, we introduce \projectname, the first learned method for estimating the optimal light source powers required for illuminating multi-color holograms.
For this purpose, we establish the first multi-color hologram dataset using synthetic images and their depth information.
We generate these synthetic images using a trending pipeline combining generative, large language, and monocular depth estimation models.
Finally, we train our learned model using our dataset and experimentally demonstrate that \projectname significantly decreases the number of steps required to optimize multi-color holograms from $>1000$ to $70$ iteration steps without compromising image quality.  
\end{abstract}


\section{Introduction}
\label{sec:intro}  
\CGH is an emerging technology for next-generation displays, including virtual reality headsets, augmented reality glasses \cite{koulieris2019near}, and 3D displays~\cite{urey2011state}.
Through \CGH, holographic displays promise to reproduce realistic images by reconstructing accurate light fields \cite{chen2022off,eybposh2020deepcgh} or perceptually accurate representations \cite{walton2022metameric,aksit2022perceptually}.
A standard holographic display comprises a phase-only \SLM and multiple light sources that helps generate a full-color image.
Typically, the phase-only \SLM plays a single-color hologram for each color channel time-sequentially.
Meanwhile, each corresponding light source for each color channel lits a presented single-color hologram.
This way, \HVS integrates each color from these holograms, and users can observe full-color \3D scenes from holographic displays.

Recently, Chen et al. \cite{chen2022auto} optimized light source powers in single-color holograms using a camera-in-the-loop approach to accurately represent color and brightness levels.
Assuming light sources operate at their peak intensities, standard single-color holograms are limited in their dynamic range, a set of brightness levels they can represent.
This dynamic range issue becomes apparent as each light source roughly operates one-third of the time when representing a full-color image.
A recent study proposes multi-color holograms~\cite{kavakli2023multicolor} for overcoming this issue.
Their work co-optimizes multi-color holograms with their corresponding powers by each light source.
While their study improves dynamic range and brightness up to $\times1.8$ than single-color holograms, the co-optimization process requires many iterations (\eg >1000 steps).
Thus, multi-color hologram optimization takes minutes, remarkably slower than a few seconds of single-color computations.

\begin{figure}[!ht]
\begin{centering}
\includegraphics[width=1.0\columnwidth]{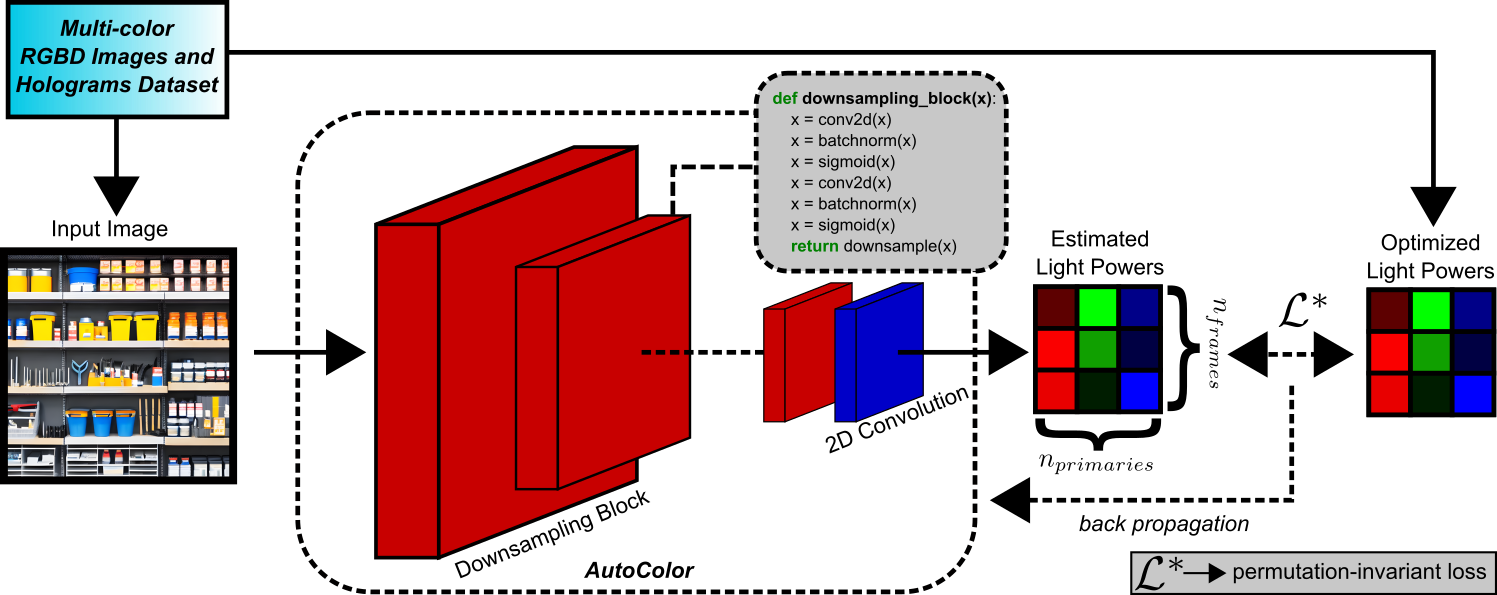}
\caption{
\projectname light power estimation network structure. 
\projectname learns to estimate powers for each light source to illuminate multi-color holograms using our multi-color hologram dataset and a permutation-invariant loss tailored for multi-color holograms.
}
\label{fig:schematic}
\end{centering}
\end{figure}

This letter proposes the first learned method, \projectname, to estimate the optimal light source powers for multi-color holograms.
\projectname reduces the multi-color optimization in the previous study~\cite{kavakli2023multicolor} from many minutes to a couple of ten seconds.
Firstly, \projectname needs a dataset to adequately train a light source powers estimation network.
Today, high-quality hologram datasets~\cite{shi2021towards,shi2022end} exist but are strictly for single-color holograms.
The multi-color hologram dataset is not readily available in the literature.
We create the first multi-color hologram dataset using synthetic but photo-realistic images and their depth information.
Secondly, we develop a \CNN architecture that consists of downsampling and convolution operations to estimate light source powers from an input image.
We train this \CNN using our multi-color hologram dataset.
Finally, we show that our \CNN significantly accelerates multi-color optimizations.
We also experimentally verified our findings in a holographic display prototype.

\section{Related Work}

The field of \CGH has witnessed significant advancements by incorporating neural networks, addressing the challenge of balancing speed and image quality. 
Liang \etal \cite{shi2021towards, shi2022end} proposed a \CGH deep learning network and a hybrid supervised+unsupervised training approach, which enables the synthesis of high-quality 3D phase-only holograms at an interactive rate.

Similarly, Choi \etal \cite{Choi2020CITL} developed a comprehensive CGH framework with a camera-in-the-loop optimization strategy. 
This innovative approach has led to the real-time generation of full-color, high-quality holographic images at 1080p resolution, a breakthrough in holographic imaging.
Furthermore, the work by Zhu \etal \cite{Zhu2023Optica} leverages Fourier basis functions in learning-based \CGH, 
demonstrating impressive performance in model generalization and quality of reconstructions.

\section{Methodology}

The schematic diagram of our learned method, \projectname, is shown in \refFig{schematic}.
To achieve \projectname, we \emph{first generate a dataset} of images with their depths, laser powers, and holograms.
We leverage a \LLM GPT-4 \cite{brown2020language} via its online interface ChatGPT to guide the generation diversity.

We use those LLM-generated prompts to develop a large dataset of images locally using text-to-image generation models, Stable Diffusion~\cite{rombach2021highresolution} (see \href{https://github.com/stability-ai/stablediffusion}{GitHub:stability-ai/stablediffusion}).
Using the publicly available weights (\href{https://huggingface.co/stabilityai/stable-diffusion-2-1-base/blob/main/v2-1_512-ema-pruned.ckpt}{v2-1$\_$512-ema-pruned.ckpt}) for the text-to-image generation model and using 12 GB memory, we generate 8865 images with $512\times512$ resolutions.
We then upsample these 8865 images to $2048\times2048$ resolutions using a \GAN based super-resolution network~\cite{wang2021realesrgan} (see \href{https://github.com/xinntao/Real-ESRGAN/}{GitHub:xinntao/Real-ESRGAN}).
The upsampling process runs locally on an RTX 3090, consuming 4.7 GB memory using the publicly available weights (\href{https://github.com/xinntao/Real-ESRGAN/releases/download/v0.1.0/RealESRGAN_x4plus.pth}{RealESRGAN$\_$x4plus.pth}).
To estimate the depth information for all the generated images at $2048\times2048$ resolution, we rely on a monocular depth information network~\cite{Ranftl2021, Ranftl2022} and their publicly available weights (\href{https://github.com/intel-isl/DPT/releases/download/1_0/dpt_hybrid-midas-501f0c75.pt}{dpt$\_$hybrid-midas-501f0c75.pt}), consuming 35 GB memory on an NVIDIA A100 on a cloud GPU cluster.
We also downscale 8865 RGBD images with $1024 \times 1024$ resolutions to reduce time and memory consumption during the hologram generation step.
Locally, we optimize multi-color holograms and their light source powers using \textit{Multicolor} optimization pipeline~\cite{kavakli2023multicolor} (see \href{https://github.com/complight/multicolor}{GitHub:complight/multicolor}).
To efficiently manage our workload, we found it imperative to employ multiple GPUs,
We convert all the RGB images to multi-color holograms at $\times1.8$ brightness, $1024 \times 1024$ resolution, and $8 \mu m$ pixel pitch.
We target three depth layers and 500 steps for each multi-color hologram, with the learning rate starting from 0.025 and decaying to 0.005 (highly aggressive to coarse learning rates).
The multi-color holograms are generated for three target planes, which are on -0.5 cm, 0 cm, and 0.5 cm with respect to our \SLM (hologram plane).
Our entire dataset generation consumes about ten days of computation using multiple GPUs.

Conventional single-color holograms rely on a field-sequential color method and use one single monochromatic light source at a time.
\HVS fuses these single-color images into a full-color as these holograms are displayed at rates well above \CFF.
Assuming that a holographic display has three monochromatic light sources, each of them is optimized by solving the following equation,
\begin{equation}
\optmSlmPhasePrimary \leftarrow \operatorname*{argmin}_{\slmPhasePrimary} \sum_{\pIndex=1}^{3} \lossFunc({\lvert e^{i\slmPhasePrimary} * \propKernel \rvert}^2, \tgtIntensity),
\label{eq:conventional_optm}
\end{equation}
where $\pIndex$ denotes the index of a color primary, $\slmPhasePrimary$ is the \SLM phase, $\optmSlmPhasePrimary$ is the optimized single-color hologram, $\propKernel$ is the wavelength-dependent light transport kernel~\cite{matsushima2009band,kavakli2022learned}, $\tgtIntensity$ is the target image intensity, $*$ denotes the convolution operation, and $\lossFunc$ denotes any valid loss function that measures the difference between the reconstruction and target. 
On the other hand, multi-color holograms use multiple monochromatic light sources simultaneously.
Assuming the total number of subframes to  $\numSubFrames=3$ like in the conventional single-color holograms, multi-color hologram generation could be formulated as
\begin{equation}
\optmSlmPhaseSubFrame, \optmLaserIntensity_{(\pIndex, \subFrameIndex)} 
\leftarrow
\operatorname*{argmin}_{\slmPhaseSubFrame, \laserIntensity_{(\pIndex, \subFrameIndex)}} \underbrace{\sum_{\pIndex=1}^{3} \left\lVert \left(\sum_{\subFrameIndex=1}^{\numSubFrames} \left\lvert \sqrt{\laserIntensity_{(\pIndex, \subFrameIndex)}} e^{i\frac{\wavelength_{\pIndex}}{\wavelength_{\pAnchor}}\slmPhaseSubFrame} * \propKernel \right\rvert^2\right) - \scale \tgtIntensity\right\rVert_2^2}_{\lossTerm_{\text{image}}},
\label{eq:holohdr_optm}
\end{equation}
where $\laserIntensity_{(\subFrameIndex, \pIndex)}$ represents the light source intensity for the $\pIndex$-th primary at the $\subFrameIndex$-th subframe, $\wavelength_\pIndex$ denotes the wavelength of the active primary, $\wavelength_{\pAnchor}$ denotes the wavelength of the anchor primary, for which the nominal value of the SLM phase is calibrated against (\eg $\wavelength_{\pAnchor} = 515~nm$ in our hardware prototype), and $\scale$ determines how bright a final image should be.
For this study, we choose $\scale = \times 1.8$, whereas single-color hologram has lower brightness values with $\scale = \times 1.0$.
Specifically, a multi-color hologram optimization seeks the optimal light source intensity, $\laserIntensity$ (a $3\times3$ matrix for light sources and 3 subframes).
Values in $\laserIntensity$ are always normalized between zero (complete switching) and one (the peak brightness level of a monochromatic light source).
If $\scale>1$, $\laserIntensity$'s rows representing subframes will sum up to a value $>1$ to match the demand of $\scale$ (code implementation at \href{https://github.com/complight/multicolor}{GitHub:complight/multicolor}).
For single-color holograms, $\laserIntensity$ is a preset value that corresponds to an identity matrix where only one monochromatic light source operates at each subframe.
Certainly, it does not meet the demand of $\scale>1$ effectively as the $\laserIntensity$'s rows sum up to one and often yields image degradation as described in \cite{kavakli2023multicolor}.
In this work, we estimate $\laserIntensity$ matrix to provide a faster convergence rate in multi-color hologram optimizations.

Leveraging the dataset and the described multi-color optimizations, we develop \projectname, a light source power estimation \CNN using PyTorch~\cite{paszke2017automatic} and Odak~\cite{kavakli2022optimizing}, where we estimate various $\laserIntensity$ from input images.
\projectname includes downsampling blocks followed by a final convolutional layer, as shown in \refFig{schematic}.
Each downsampling block has a cascade of 2D convolution layers with a kernel size of three and a channel size of twenty-four.
Each convolutional layer in a downsampling block is followed by batch normalization and nonlinear activation function.
The last layer of each block is a downsampling operation.
Our \CNN contains three downsampling blocks, starting from original image resolution to downsampling to $100\times100$, $10\times10$, and $3\times3$ in stages.
Using an RTX 2080 Ti, we train our \CNN for 40 epochs, starting with a learning rate of 0.002, and declining to 0.0005.
Our training with an Adam solver includes all the images and optimized laser powers of our multi-color holograms from our dataset. The ground truth laser powers of the images are generated from \cite{kavakli2023multicolor}.
We include regularization terms for various considerations.
(1) The rows of $\laserIntensity$ provide light source power for individual frames.
Their ordering, however, shall not alter the reconstructed imagery.
Therefore, the predicted laser light source power is regularized for order invariance.
(2) The estimated values shall be bounded with a physically plausible range $[0,1]$.
The regularization loss term is constructed as follows, 
\begin{equation}
\lossTerm = \underbrace{\operatorname*{argmin}_{m \in \{k!\} } \left\lVert ^m \laserIntensity_{est} - \laserIntensity_{opt} \right\rVert_2^2}_{\lossTerm_{invariant}} + \underbrace{\sum \left\lvert \laserIntensity_{est} < 0 \right\rvert + \sum \left\lvert \laserIntensity_{est} > 1 - 1 \right\rvert }_{\lossTerm_{normalize}}.
\end{equation}
Here, $\lossTerm_{invariant}$ and $\lossTerm_{normalize}$ are loss functions to encourage frame-invariance and bounding to $[0,1]$.
Meanwhile, $k$ represents frame permutations ($m$ for each of them), $\laserIntensity_{est}$ represents light source power estimation, and $laserIntensity_{opt}$ represents the corresponding optimized power from our dataset.
Note that the way we choose the minimum loss among permutations is a vital component of our training, it is similar to permutation-invariant loss functions in speech separation literature~\cite{yu2017permutation}.
After the training converges, we use the estimated light source powers from the \CNN and optimize multi-color holograms using~\cite{kavakli2023multicolor}.
We observe that the paradigm significantly improved computational efficiency by reducing the optimization to 70 steps, compared with the 1000 steps from prior work \cite{kavakli2023multicolor}.

\section{Evaluation}

\begin{figure*}[!ht]
  \begin{centering}
  \includegraphics[width=\textwidth]{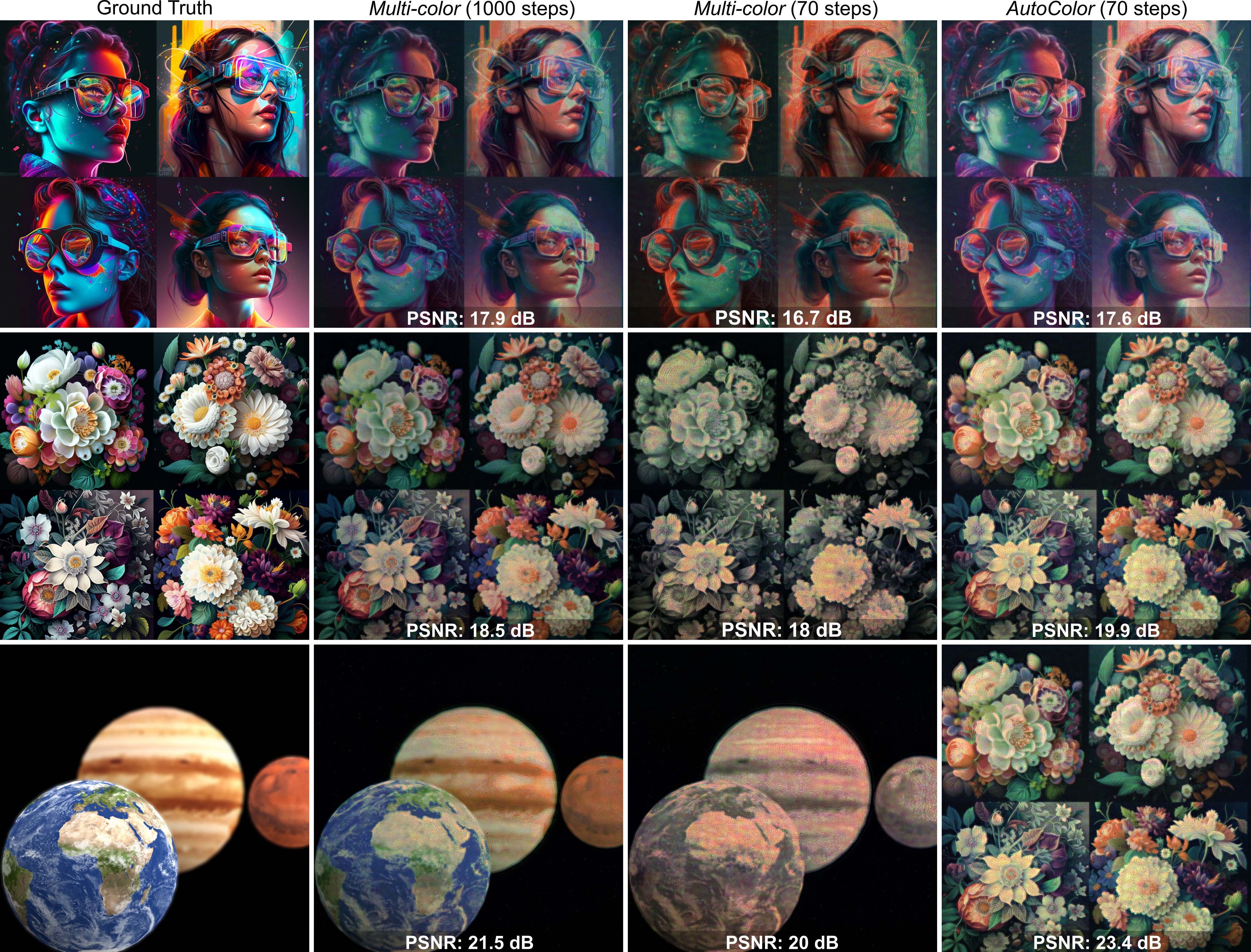}
  \caption{
    Photographs showing \projectname generating x1.8 times brighter images in lesser steps than \textit{Multicolor}. 
    Experimental results show that \projectname achieves high-fidelity visuals using only 70 steps, whereas \textit{Multicolor} requires 1000 steps for similar quality and fails to produce correct color information in 70 steps. 
    Images optimized with 70 steps using \projectname provides similar quantitative image metrics (see the inset numbers) compared with the images generated with 1000 steps in \textit{Multicolor}.
    AutoColor applies to both 2D images (first and second row) and 3D images (third row).
    (Source link: \href{https://github.com/complight/images}{\textbf{Github:complight/image}}, 80 ms exposure).
  }
  \label{fig:results_figure}
  \end{centering}

\end{figure*}

As a \emph{quantitative evalutation}, we developed a holographic display hardware prototype following~\cite{kavakli2023multicolor}.
Our prototype uses a LASOS MCS4 multi-wavelength laser (473, 515, and 639 nm) with controlled power levels from two ESP32-WROOM-32D modules. 
A pinhole aperture, Thorlabs SM1D12, was placed in front of the fiber to limit the numerical aperture of the diverging beams.
Thorlabs LPVISE100-A linear polarizer placed after the pinhole aperture allows a polarization state aligned with the SLMs fast axis for light beams. 
These linearly polarized beams get modulated with our phase-only SLM, Holoeye Pluto-VIS ($1080 \times 1920$ resolution and $8 \mu m$ pixel pitch), and arrive at a 4f imaging system composed of two 50 mm focal length achromatic doublet lenses and a pinhole aperture (Thorlabs AC254-050-A and SM1D12).
We capture the image reconstructions with a Ximea MC245CG-SY camera, located on an X-stage (Thorlabs PT1/M range: 0-25 mm, precision: 0.01 mm).
The prototype is configured as an off-axis imaging system. A linear grating term was applied to phase holograms to generate images at the half-diffraction order location.
We calculate the linearly grated phase hologram, $O_h'$,
\begin{equation}
  O_h'(x,y) =
  \begin{cases}
            e^{-j(\phi(x,y) + \pi)} & \text{if $y=$ odd} \\
            e^{-j\phi(x,y)}         & \text{if $y=$ even} \\
  \end{cases}
\end{equation}
where $\phi$ represents the original phase values of $O_h$ at the $x$ and $y$ pixel locations.  
Our quantitative evaulation is summarized in \refFig{results_figure}.

\section{Conclusion}

In conclusion, we show that multi-color hologram optimizations could be achieved with significantly fewer steps.
To this aim, we develop a light power estimation network, \projectname, powered by our first multi-color hologram generation pipeline and dataset.
The findings are further validated with an experimental analysis.
We hope \projectname to pave the way toward an exciting research frontier for future holograms with wide dynamic ranges at interactive rates.

In the future, we envision to advance the research on adaptability and energy efficiency. 
Our \CNN is trained for a fixed multi-color hologram generation routine, where $\scale=\times 1.8$.
The architecture may be further improved such that $scale$ conditions the estimated light source powers.
This way, the other $\scale$ values could be requested from the \CNN.
To do so, the multi-color hologram dataset shall be extended to include varied $\scale$ values.
Similarly, the \CNN structure estimates optimal light source powers, instead of the minimal energy consumption.
In fact, power savings could be realized if a gaze tracker is introduced in the system and target colors are chosen following \HVS characteristics in foveal and peripheral regions similar to \cite{duinkharjav2022color,chakravarthula2021gaze}.

\medskip

\noindent \textbf{Funding.} Kaan Akşit, Koray Kavaklı and Yicheng Zhan are supported by the Royal Society's RGS/R2/212229 - Research Grants 2021 Round 2 and  Meta Reality Labs inclusive rendering initiative 2022. 
Hakan Urey is supported by the European Innovation Council's HORIZON-EIC-2021-TRANSITION-CHALLENGES program Grant Number 101057672 and Tübitak’s 2247-A National Lead Researchers Program, Project Number 120C145.
Qi Sun is partially supported by the National Science Foundation (NSF) research grants \#2225861 and \#2232817.

\medskip

\noindent \textbf{Acknowledgements.} 
The authors thank the funders and anonymous reviewers for their feedback.
The authors would also like to thank Liang Shi for fruitful discussions.

\medskip

\noindent \textbf{Disclosures.}
The authors declare no conflicts of interest.

\medskip

\noindent \textbf{Data availability.}   
All data needed to evaluate the conclusions in the manuscript are provided in the manuscript.
Source code and dataset can be downloaded from \href{https://github.com/complight/autocolor}{GitHub:complight/autocolor}.
Additional data related to this paper may be kindly requested from the authors.

\noindent \textbf{Supplemental document.}
 See Supplement 1 for supporting content.

\medskip
\noindent
\bibliography{references} 
\bibliographystyle{spiebib} 

\end{document}